# Medical Specialty Prediction from Bangla Patient Narratives: A Comparative Study of ML, DL, Transformers, and LLMs

*Tasnin Tabassum Ali*[1,*], *Faisal Ahmed*[2], *& Muhammad Anwarul Azim*[1]

[1]Department of Computer Science and Engineering, University of Chittagong, Chattogram, Bangladesh
[2]Universitat Rovira i Virgili, Tarragona, Spain; Email: *faisalcsecubd@gmail.com*

[*] Corresponding Author: *Tasnin Tabassum Ali; 20701046@std.cu.ac.bd*; *Muhammad Anwarul Azim: azim@cu.ac.bd*

**Abstract**

Timely access to appropriate medical specialists remains a bottleneck in Bangladesh's healthcare system, as patients describe symptoms in informal Bangla rather than clinical terms, leading to ineffective referrals. This paper compares four model families for predicting medical specialties from Bangla patient narratives: six machine learning (ML) models using hybrid TF-IDF features, four deep learning (DL) models using Word2Vec embeddings trained on the corpus, four fine-tuned transformers, and five large language models (LLMs) evaluated zero-shot. From a corpus of 3798 patient records covering 20 specialties, we retained the five with the most records (Medicine, Gynecology, Dermatology, Pediatrics, Gastroenterology) and split the resulting 2730 records 80:20 into training and test sets shared by all families. The best transformer was XLM-RoBERTa (86.26% accuracy, 86.21% macro-F1), the best ML model was SVM (84.98%, 84.92%), the best DL model was Bidirectional GRU (80.77%, 80.83%), and the best LLM was Gemini (90.29%, 90.15%). Holm-corrected paired McNemar tests show SVM significantly outperforming Bi-LSTM (adjusted $p = 0.004$), whereas the Gemini-versus-Grok gap does not survive correction (adjusted $p = 0.104$), and neither the strongest transformer versus the weakest LLM (Grok) nor SVM versus the strongest transformer reaches significance. A tuned classical model is thus statistically indistinguishable from the strongest transformer, so accuracy rankings alone should not be read as evidence of one family's superiority. The findings provide a comparative baseline for Bangla medical text classification.

## 1. Introduction

### *1.1. Background and Motivation*

Bangladesh's healthcare-delivery infrastructure places a disproportionate burden on patients to self-triage to access an appropriate specialist. Most initial interactions between patients and the health system now take place informally - through posts in medical-advice social-media groups, or free-text descriptions at hospital reception desks - due to long queues, uneven distribution of specialists between urban and rural facilities, and low health literacy. These narratives are written in colloquial Bangla rather than clinical terms, which causes loss of clinical time and delays treatment when patients are mis-referred to the wrong specialist.

One NLP-driven solution is to predict the best-matching medical specialty from a brief patient description, which could guide automated triage or referral. NLP has proven effective for referral and emergency-department triage (Wang et al., 2023; Sánchez-Salmerón et al., 2022), although most specialty-prediction work uses English-language data (Kim et al., 2023). Bangla medical-text classification has relied mainly on classical ML and DL (Ahnaf et al., 2021), and recent work has begun to extend it to transformer-based medical entity recognition (Aurpa et al., 2026). Fine-tuned transformers and general-purpose LLMs, the latter able to classify zero-shot without task-specific training, are now widely used for text classification, yet no study compares these generations of methods on Bangla medical narratives: prior work typically evaluates one model family at a time, on different models, datasets, or languages. It therefore remains unclear whether added complexity (recurrence, attention, or scale) brings a consistent advantage on the same task. This study addresses that question and contributes a reproducible cross-family benchmark relevant to equitable, low-resource healthcare NLP.

### *1.2. Literature Review*

Classical ML remains a strong candidate for medical-text classification. Hussain Almuhana and Abbas (2022) found a Multi-Layer Perceptron the most accurate of several classical classifiers on a nine-specialty medical-report corpus, and Faris et al. (2020), the most similar prior study, reached 85% accuracy on 15,000 Arabic medical questions across 15 specialties. In deep learning, Kim et al. (2023) fine-tuned BERT for 27-way specialty prediction on patient-authored text, Lee et al. (2021) used LSTM variants in a symptom-triage chatbot, Alomari et al. (2023) addressed class imbalance in Arabic telemedicine questions with SMOTE-augmented DNNs, and Lee et al. (2025) combined BERT with entity information for specialty prediction from clinical questionnaires. Ahnaf et al. (2021), the closest work to our DL track, found LSTM, CNN, and GRU models outperforming Naive Bayes and SVM on Bangla medical text. Mao et al. (2023) also studied automatic specialty classification from patient-described symptoms using a hybrid neural model on a large Chinese-language dataset, providing a useful cross-lingual comparison.

Our transformer track uses four pretrained encoders: BanglaBERT (Bhattacharjee et al., 2022), an ELECTRA-style model pretrained for Bangla; Bangla-BERT-base, an independently pretrained BERT-base checkpoint; XLM-RoBERTa (Conneau et al., 2020), trained on 100 languages; and multilingual BERT (Devlin et al., 2019), whose EHR-pretrained descendant Med-BERT (Rasmy et al., 2021) shows the architecture's usefulness for clinical prediction. Domain-specific pretraining can help clinical NLP but is not automatic: Yao et al. (2019) gained from domain-specific BERT pretraining on Traditional Chinese Medicine records, whereas Gao et al. (2021) reported limitations of BERT-based models on long clinical documents relative to simpler baselines.

Evidence on LLMs in clinical settings is mixed. Surya Kathirvel and Mookiah (2025) found Naive Bayes outperforming BERT-family classifiers on medical-transcription specialty classification, Brown et al. (2025) found non-fine-tuned GPT-3.5/GPT-4 less effective than locally trained ML on structured clinical prediction, and Naliyatthaliyazchayil et al. (2025) reported only moderate zero-shot success on medical coding and readmission risk. In contrast, Zhu et al. (2026) found recent zero-shot LLMs increasingly outperforming fine-tuned BERT-family models on unstructured clinical notes, the modality studied here, and a systematic review of 65 studies reported LLMs consistently outperforming traditional ML in healthcare text classification (Sakai & Lam, 2026). Yang et al. (2022) showed that clinical language models improve with scale, while Thirunavukarasu et al. (2023) caution about reliability and interpretability.

Several of these studies predict specialty within a single model family (Ahnaf et al., 2021; Faris et al., 2020; Kim et al., 2023; Lee et al., 2025), but none places classical ML, recurrent DL, fine-tuned transformers, and zero-shot LLMs on one Bangla corpus and one frozen test split. We therefore hold the corpus, the label set, and the test partition fixed across all four families, so that performance differences can be attributed to model family rather than to differences in data or evaluation protocol.

### *1.3. Contribution*

We compare 19 models in four families on one Bangla specialty-prediction task: six classical ML models (Naive Bayes, Logistic Regression, Decision Tree, KNN, SVM, Random Forest) on hybrid word- and character-level TF-IDF features; four recurrent DL models (LSTM, Bi-LSTM, GRU, Bi-GRU) on Word2Vec embeddings learned from our corpus; four fine-tuned transformers (BanglaBERT, Bangla-BERT-base, XLM-RoBERTa, mBERT); and five general-purpose LLMs (ChatGPT, Gemini, Grok, Claude, DeepSeek) prompted zero-shot. LLM predictions are matched to gold labels through a persistent row identifier. Beyond accuracy and macro-F1, we add Wilson confidence intervals, a five-seed robustness check of the ML track, a bootstrap re-analysis, and Holm-corrected paired McNemar tests.

### *1.4. Theoretical Framework*

The four families correspond to successive paradigms of text representation: sparse lexical statistics (TF-IDF) in classical ML, corpus-specific embeddings (Word2Vec) in recurrent DL, pretrained contextual representations in transformers, and in-context reasoning without parameter updates in zero-shot LLMs. Each step may add representational power but may also mismatch informal Bangla narratives. RQ1 and RQ2 follow this progression; RQ3 asks whether the same classes are hardest under all four paradigms, which would point to a property of the data.

### *1.5. Research Questions*

The study addresses three research questions. **RQ1:** Do fine-tuned transformers outperform classical ML and recurrent DL baselines trained on the same narratives? **RQ2:** Can general-purpose, zero-shot LLMs match or exceed models trained on the target distribution? **RQ3:** Which specialties are hardest to predict, and what does the resulting confusion reveal about the structure of the triage problem?

## 2. Materials & Methods

### *2.1. Dataset*

This is a quantitative, comparative benchmark study in which the unit of analysis is a single patient narrative. The corpus contains 3,798 Bangla patient-problem narratives, annotated for medical specialty by majority vote of three MBBS-qualified physicians. We keep the five best-represented specialties, Medicine (645), Gynecology (591), Dermatology (584), Pediatrics (504), and Gastroenterology (406), giving 2,730 labeled records with a roughly 1.6:1 imbalance between the largest and smallest class. A single stratified 80/20 split (seed 2025) produced a 2,184-record training partition and a 546-record test partition (Medicine 129, Gynecology 118, Dermatology 117, Pediatrics 101, Gastroenterology 81), frozen and reused identically by all four model families.

The corpus was developed earlier by students, under the second author's guidance, for their academic theses and projects (see Acknowledgements) and is reused here with the second author's permission. It combines in-person field interviews with narratives from online medical-advice forums and spans twenty raw specialties; the fifteen specialties with too few records for reliable training and evaluation are excluded.

#### *Data Privacy and Ethical Considerations*

Permission to use the dataset was obtained from the relevant data sources. Patient names and other direct personal identifiers were removed before analysis to protect privacy. Three MBBS-qualified physicians reviewed the symptom descriptions for clinical relevance and assigned specialty labels through majority

voting. For LLM-based evaluation, de-identified symptom descriptions were submitted to five external LLM services. These measures were intended to reduce privacy risks; however, de-identification does not eliminate the possibility of re-identification or independently establish authorization for secondary research use and external processing. Formal institutional ethics approval was not obtained for the original data collection, and individual consent forms were not collected. Accordingly, the institutional basis for the original collection and the scope of permission for subsequent research use and third-party processing remain to be clarified with the relevant institutional authority. No claim of ethics approval or exemption is made.

Fig. 1 summarizes the overall workflow.

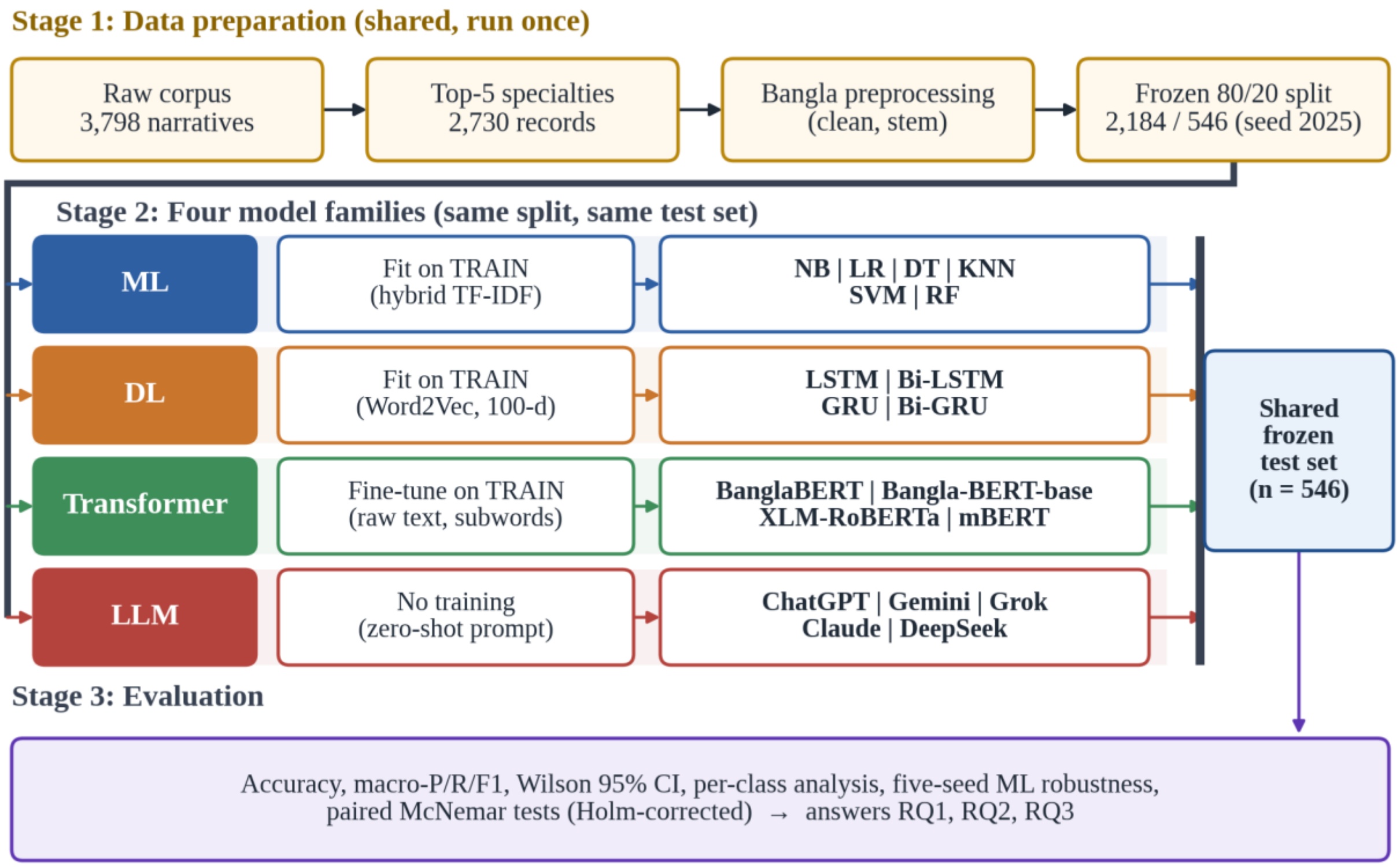

**Figure 1.** Overall workflow of Bangla medical specialty prediction; all models are then evaluated with accuracy, macro-F1, 95% CIs, and Holm-corrected McNemar tests.

Source: Authors' analysis.

### *2.2. Pre-processing*

Bangla text was Unicode-normalized, stripped of punctuation and emoji, filtered with a Bangla stop-word list (bnlp_toolkit's BengaliCorpus), and stemmed with a rule-based stemmer (bnltk's BanglaStemmer). Two representations are kept: a whitespace-joined clean-text string for the ML track, and token lists used to train a Word2Vec embedding (100-d, CBOW, window 5, 6,709-word types) for the DL track. Transformers are fine-tuned on raw, unstemmed text because subword tokenization already captures Bangla's inflectional morphology. Training narratives are short relative to the padded length used downstream (mean 17.6, median 13, 95th percentile 48 tokens, against a maximum sequence length of 100), so the recurrent embedding layer uses masking to keep padding from diluting the learned representation.

### 2.3. *System Architecture and Four Model Tracks*

Fig. 1 shows the end-to-end pipeline. The ML track fits six classical classifiers on a hybrid TF-IDF space combining word (1,2)-gram and character (3,5)-gram vectorizers (15,000 features each; 29,567 dimensions in total). TF-IDF was preferred to averaged Word2Vec features because comparative evidence shows it matching or exceeding them with linear classifiers on similarly sized corpora (Cahyani & Patasik, 2021). The DL track trains LSTM, Bi-LSTM, GRU, and Bi-GRU models on Word2Vec embeddings learned from the training narratives and updated during training, with an identical downstream head, Adam optimizer, and early stopping on validation accuracy. The transformer track fine-tunes four pretrained checkpoints on raw text with a standard sequence-classification head. The LLM track applies five general-purpose models without fine-tuning, each receiving the same fixed prompt (closed five-way label set and output format) on blinded batches of the frozen test partition; predictions are re-joined to gold labels by row identifier rather than row position, guarding against silent reordering or dropped rows (Section 2.3.1).

#### 2.3.1. *LLM Track: Model Versions and Access Configuration*

All five models, ChatGPT (GPT-5.6 "Luna"), Gemini (Gemini 3.6 Flash), Grok (Grok 4.5), Claude (Claude Sonnet 5, medium reasoning effort), and DeepSeek (DeepSeek-V3), were accessed on August 31, 2026 through each provider's chat interface, which exposes no decoding controls, so provider defaults applied. The 546 test narratives were split into 11 blinded batches of up to 50 rows, each submitted manually to each model. The fixed prompt asked each model to classify every narrative into one of five integer-coded specialties (0 = Medicine, 1 = Gynecology, 2 = Dermatology, 3 = Pediatrics, 4 = Gastroenterology), copy each row identifier unchanged, give a best single guess when unsure, and return only a two-column CSV (row_id, label); the full prompt is provided with the code. Claude returned every batch directly as the specified two-column CSV; the other four models answered in other formats, so their responses were converted to CSV in a separate Claude session and checked manually against the raw output to confirm that row identifiers and labels were preserved exactly. Per-model files were then combined and re-joined to gold labels by row identifier. No output was malformed, empty, or refused, so none was excluded or corrected. Because providers update models without versioning, exact replication cannot be guaranteed beyond the access date.

### 2.4. *Evaluation and Statistical Testing*

Every model is scored once on the same 546-row test partition using accuracy and macro-averaged precision, recall, and F1. We also report 95% Wilson score intervals (Wilson, 1927) for each model's accuracy. For the cross-family contrasts most relevant to the research questions we run McNemar's test (McNemar, 1947) on each model's actual per-row predictions, which uses row-level agreement: the exact binomial form when fewer than 25 discordant pairs exist, and the continuity-corrected chi-square form otherwise. Because four pairwise tests share the same 546 rows, Holm's step-down procedure controls the family-wise error rate, and a 5,000-sample bootstrap of paired accuracy differences serves as a secondary check. Reliability is examined through a five-seed robustness analysis of the ML track (Section 3.2); validity of the family comparison rests on scoring every model on identical test rows and testing paired row-level predictions rather than independent summary statistics.

### 2.5. *Reproducibility Details*

**Software environment.** All experiments ran in Google Colab (Python, NVIDIA T4 GPU) with scikit-learn 1.9.0, gensim 4.4.0, bnlp_toolkit 4.4.1, bnltk 0.7.8, TensorFlow/Keras, and Hugging Face Transformers (the last two not pinned). Seed 2025 was fixed for NumPy, Python's random module, the train/test split, all ML models, Keras, Word2Vec, and the Trainer; the transformers' classification-head initialization was not seeded separately, so reruns may differ slightly. Table 1 lists the model settings.

**Validation protocol.** ML classifiers were fit on the full 2,184-row training partition and scored once on the 546-row test partition, with no held-out validation split. The DL track held out 10% of the training rows (about 218) for early stopping (patience 4, best weights restored). The transformer track used a fixed five-epoch schedule with no checkpoint selection; the test partition was evaluated after each epoch for monitoring only and did not influence the final parameters, but a dedicated validation split would be preferable and is planned as future work. Code and trained models can be requested from the corresponding author; access to the dataset is subject to the permission of its developers (see Acknowledgements).

**Table 1.** Hyperparameters and model settings.

| Component | Setting |
|---|---|
| TF-IDF (ML) | Word (1,2)-grams + char_wb (3,5)-grams, 15,000 features each, sublinear_tf=True; 29,567 dimensions combined |
| Classical models | MultinomialNB (default priors); LogisticRegression (max_iter=3000, class_weight=balanced); DecisionTree (defaults); KNN (k=9, cosine); LinearSVC (class_weight=balanced, max_iter=8000); RandomForest (400 trees); random_state=2025 |
| Word2Vec (DL) | 100-d CBOW (sg=0), window 5, min_count=1, 6,709 word types plus padding index; embedding layer initialized from it and trainable; mask_zero=True; MAXLEN=100 |
| Recurrent models | Two stacked layers (128→64 units; uni- or bidirectional) → Dense(128, ReLU) → Dropout(0.4) → Softmax(5); Adam (lr=1e-3); batch 32; up to 25 epochs; early stopping (patience 4, 10% validation split) |
| Transformers | Learning rate 2e-5; batch size 16 (train) / 32 (eval); 5 epochs; weight decay 0.01; max sequence length 128; seed 2025; no checkpoint selection |

Source: Authors' analysis.

## 3. Results and Discussions

### *3.1. Findings*

Table 2 reports accuracy, macro-averaged precision, recall, and F1, and the 95% Wilson interval for all 19 models, ranked by accuracy.

By best model per family, LLM (90.29%) > transformer (86.26%) > ML (84.98%) > DL (80.77%); every LLM outscored every trained model on raw accuracy, and macro-F1 reorders no family. As Section 3.3 shows, however, most of these gaps are statistically indistinguishable once multiple comparisons are corrected for, so the ordering is descriptive, not evidence of family-level superiority. XLM-RoBERTa leads BanglaBERT by only 1.46 points, within the noise of a 546-row partition, so multilingual pretraining may substitute for language-specific pretraining at this scale. SVM and Logistic Regression exceed KNN and Decision Tree by 13.74 to 19.05 points, consistent with linear models exploiting the sparse TF-IDF space. Bidirectional models beat their unidirectional counterparts, suggesting that right-context aids disambiguation.

Accuracy alone can also mislead. Random Forest and Naive Bayes tie at 78.57%, yet Random Forest's macro-F1 (78.54%) exceeds Naive Bayes' (78.06%) by nearly half a point, because Naive Bayes recalls only 62% of the smallest class, Gastroenterology, against 75% for Random Forest (notebook class reports). This is why we treat macro-F1 as the primary metric for close calls.

### *3.2. Five-Seed Robustness (ML track)*

To check whether the ML ranking in Table 2 is an artifact of the single frozen split, all six classical models were rerun over five seeds (2025, 7, 42, 123, 2026), each changing the train/test split and, where relevant, model-internal randomness. Mean accuracy (± SD) was: SVM 83.99% ± 1.59 (macro-F1 83.94% ± 1.74), Logistic Regression 83.19% ± 2.00 (83.07% ± 2.09), Random Forest 79.30% ± 1.19 (79.06% ± 1.15), Naive Bayes 77.88% ± 1.37 (76.79% ± 1.49), KNN 71.72% ± 2.62 (71.51% ± 2.77), and Decision Tree 64.69% ± 0.96 (64.41% ± 1.11). SVM had the highest mean accuracy and macro-F1, but its 0.8-point margin over Logistic Regression is smaller than either model's SD, so the two are comparable, while both clearly exceed the other four. This rerun used lighter preprocessing and fewer solver iterations, so it supports ranking stability, not exact reproduction of Table 2.

**Table 2.** Full results for all 19 models on the 546 test rows (values in %; Acc. = accuracy, Prec. = macro precision, Rec. = macro recall, F1 = macro-F1, CI = confidence interval).

| Family | Model | Acc. | Prec. | Rec. | F1 | 95% CI (Acc.) |
|---|---|---|---|---|---|---|
| LLM | Gemini | 90.29 | 89.93 | 90.74 | 90.15 | [87.52, 92.50] |
| LLM | DeepSeek | 89.19 | 88.92 | 89.49 | 89.07 | [86.31, 91.53] |
| LLM | Claude | 89.19 | 88.83 | 89.51 | 89.03 | [86.31, 91.53] |
| LLM | ChatGPT | 89.01 | 88.71 | 89.29 | 88.77 | [86.11, 91.37] |
| LLM | Grok | 88.28 | 87.94 | 88.47 | 88.09 | [85.31, 90.71] |
| Transformer | XLM-RoBERTa | 86.26 | 86.16 | 86.37 | 86.21 | [83.12, 88.90] |
| ML | SVM | 84.98 | 84.90 | 85.12 | 84.92 | [81.74, 87.73] |
| Transformer | BanglaBERT | 84.80 | 84.72 | 84.79 | 84.67 | [81.54, 87.57] |
| ML | Logistic Regression | 83.15 | 83.11 | 83.51 | 83.13 | [79.78, 86.06] |
| Transformer | Bangla-BERT-base | 82.42 | 82.56 | 82.64 | 82.57 | [79.00, 85.38] |
| Transformer | mBERT | 81.32 | 81.32 | 81.62 | 81.42 | [77.83, 84.37] |
| DL | Bi-GRU | 80.77 | 81.28 | 80.53 | 80.83 | [77.25, 83.86] |
| DL | Bi-LSTM | 79.67 | 81.16 | 78.85 | 79.44 | [76.09, 82.83] |
| ML | Random Forest | 78.57 | 79.06 | 78.79 | 78.54 | [74.94, 81.81] |
| ML | Naive Bayes | 78.57 | 79.04 | 77.74 | 78.06 | [74.94, 81.81] |
| DL | LSTM | 78.39 | 80.76 | 78.31 | 78.62 | [74.74, 81.64] |
| DL | GRU | 77.84 | 78.77 | 76.85 | 77.54 | [74.17, 81.12] |
| ML | KNN | 69.41 | 71.38 | 68.02 | 68.77 | [65.42, 73.13] |
| ML | Decision Tree | 65.93 | 65.75 | 65.40 | 65.53 | [61.86, 69.79] |

Source: Authors' analysis.

### *3.3. Statistical Significance*

Confidence intervals alone understate how distinguishable two models are when scored on the same rows, because they ignore row-level agreement. Table 3 reports paired McNemar tests on four pairs chosen to cover each cross-family contrast relevant to RQ1 and RQ2 and the spread within the LLM family, using each model's actual per-row predictions.

**Table 3.** Paired McNemar's test on key model pairs (n = 546). a = rows only model 1 classifies correctly; b = rows only model 2 classifies correctly.

| Pair | Disagreements (a/b) | Raw p | Holm-adjusted p | Result |
|---|---|---|---|---|
| SVM vs. Bi-LSTM | 51 / 22 | 0.0010 | 0.0040 | Significant |
| Gemini vs. Grok | 17 / 6 | 0.0347 | 0.1041 | Not significant |
| XLM-RoBERTa vs. Grok | 28 / 39 | 0.2218 | 0.4436 | Not significant |
| SVM vs. XLM-RoBERTa | 31 / 38 | 0.4701 | 0.4701 | Not significant |

Source: Authors' analysis.

Two findings stand out. First, although the confidence intervals of Gemini and Grok overlap, their discordant counts (17 vs. 6) give a nominally significant raw difference (p = 0.035) that does not survive Holm correction (adjusted p = 0.104), which illustrates why correction matters when several pairs share test rows. Second, SVM is not significantly different from XLM-RoBERTa (adjusted p = 0.470) despite a 1.28-point accuracy gap, so a tuned classical model is a statistically defensible and far cheaper alternative to a fine-tuned transformer on this dataset. By contrast, SVM significantly outperforms Bi-LSTM, a strong recurrent model (adjusted p = 0.004), so the ML-versus-DL gap in Table 2 is unlikely to be sampling noise; Bi-GRU, the best DL model, was not paired with SVM. A 5,000-sample bootstrap of the same paired differences agrees with three of the four McNemar conclusions; for Gemini versus Grok the unadjusted bootstrap interval excludes zero (mean difference +0.020, 95% CI [+0.004, +0.039]) because it was not adjusted for multiple testing.

### *3.4. Baseline Comparison*

As reference points, uniform random guessing yields 20.01% accuracy (macro-F1 19.85%) and stratified random guessing 20.48% (19.97%), both Monte Carlo estimates from the test class distribution, whereas always predicting the majority class (Medicine) yields 23.63% accuracy but only 7.64% macro-F1, because macro-F1 penalizes ignoring minority classes. All 19 models exceed both random baselines by more than 45 accuracy points; the weakest model (Decision Tree, 65.93% accuracy, 65.53% macro-F1) and the strongest (Gemini, 90.29%, 90.15%) bracket the observed range.

### *3.5. Error Analysis*

Table 4 gives per-class results averaged over all 19 models, plus mean recall by model family, computed from each model's per-class report on the frozen test set. Medicine is the hardest class overall (mean F1 75.0%, mean recall 74.9%), followed closely by Gastroenterology (mean F1 76.2%); Gynecology is the easiest by a wide margin (mean F1 89.7%, mean precision 93.5%).

Medicine has the lowest recall in the transformer and LLM families and the second-lowest in ML, behind Gastroenterology; in DL, Gastroenterology is lowest (71.3%) and Pediatrics (75.5%) falls just below Medicine (76.0%). LLMs raise recall substantially for four classes, but not for Medicine (77.2% vs. 76.9% for transformers). Medicine's difficulty plausibly reflects its role as a catch-all category for non-specific complaints overlapping the other four specialties. Gastroenterology's difficulty looks more like a training-support effect (406 of 2,730 records, the smallest class), since it is weakest mainly for the two families trained from scratch (ML, DL). Both explanations are untested hypotheses.

**Table 4.** Per-class precision, recall, and F1 averaged over all 19 models, and mean recall by model family (values in %).

| Class | n | Mean Prec. | Mean Rec. | Mean F1 | Recall: ML | Recall: DL | Recall: Transf. | Recall: LLM |
|---|---|---|---|---|---|---|---|---|
| Medicine | 129 | 75.6 | 74.9 | 75.0 | 70.9 | 76.0 | 76.9 | 77.2 |
| Gastroenterology | 81 | 75.9 | 77.1 | 76.2 | 69.1 | 71.3 | 82.1 | 87.4 |
| Pediatrics | 101 | 84.4 | 84.1 | 83.8 | 77.4 | 75.5 | 88.1 | 95.6 |
| Dermatology | 117 | 82.3 | 87.0 | 84.5 | 83.8 | 85.5 | 85.5 | 93.3 |
| Gynecology | 118 | 93.5 | 86.4 | 89.7 | 80.9 | 85.0 | 86.7 | 93.9 |

Source: Authors' analysis.

### *3.6. Answering the Research Questions*

**RQ1 (transformers vs. ML/DL):** Partially. All four transformers numerically outperform every DL model, and SVM significantly outperforms Bi-LSTM (adjusted $p = 0.004$). However, only XLM-RoBERTa exceeds the best ML model, and not significantly (adjusted $p = 0.470$). The ML and DL tracks also differ in classifier and input representation (TF-IDF vs. Word2Vec), so this is a pipeline-level rather than an architectural comparison; an ablation (e.g., a linear classifier over averaged Word2Vec features) is needed to isolate architecture.

**RQ2 (zero-shot LLMs vs. trained models):** Descriptively yes, statistically limited. All five LLMs numerically outperform all 14 trained models, and the weakest LLM (Grok, 88.28%) exceeds the strongest transformer by two points, but that gap is not significant after Holm correction (adjusted $p = 0.444$). The best LLM (Gemini) was not part of the corrected test set, so we do not claim equivalence between LLMs and trained models in general. This agrees with Zhu et al. (2026) and Sakai and Lam (2026), and contrasts with Brown et al. (2025), whose weaker LLM results concerned structured EHR data, a difference plausibly due to data modality. Our finding is scoped to text-classification-style clinical NLP and is descriptive rather than statistically confirmed.

**RQ3 (hardest classes):** Medicine and Gastroenterology are the hardest classes overall and the two lowest-recall classes in the ML, transformer, and LLM families (Table 4); in DL, Pediatrics also falls below Medicine. Section 3.5 discusses likely causes; this argues for targeted data collection or class-weighted training in deployment.

### *3.7. Practical Considerations and Limitations*

LLMs achieve the best accuracy here but add per-query cost, latency, and third-party exposure of patient text, which a locally hosted SVM or transformer avoids. A tuned SVM comes within 5.3 points of the best LLM with no GPU or network dependency, which suits low-connectivity settings; training time, latency, and memory were not measured.

Several limitations apply. First, all comparisons rest on one stratified test partition; intervals and McNemar tests do not replace repeated-split or k-fold evaluation. Second, the ML-versus-DL comparison confounds classifier with representation. Third, the four McNemar pairs were selected (the best LLM, Gemini, was not paired with any trained model), so other pairs are exploratory. Fourth, the LLM track is a single snapshot from late August 2026 obtained through chat interfaces with default decoding and manual reformatting of four models' outputs; prompt sensitivity and pretraining-data contamination were not assessed, although overlap is plausible for publicly posted forum text. Fifth, the transformer track monitored the test set each epoch and lacks a dedicated validation split. Finally, the corpus covers five specialties from one region, and its data-use and ethical status requires clarification with the relevant institutional authority; generalization is untested. The forced-choice classifier (five of twenty raw categories, with no abstain option) should serve only as clinician-in-the-loop decision support, not as an autonomous router.

## 4. Conclusion

This study compared classical ML, recurrent DL, fine-tuned transformers, and zero-shot LLMs on 546 held-out Bangla patient narratives spanning five specialties. Gemini scored highest (90.29% accuracy, 90.15% macro-F1), followed by DeepSeek and Claude (89.19%), ChatGPT (89.01%), and Grok (88.28%); XLM-RoBERTa led the transformers (86.26%), SVM the classical models (84.98%), and Bi-GRU the recurrent models (80.77%). After Holm correction only SVM versus Bi-LSTM remained significant (adjusted p = 0.004); Gemini versus Grok, XLM-RoBERTa versus Grok, and SVM versus XLM-RoBERTa did not (adjusted p = 0.104, 0.444, 0.470). The main contribution is a reproducible single-corpus benchmark showing that a tuned classical model is statistically indistinguishable from the best fine-tuned transformer, and that the weakest LLM does not significantly outperform that transformer, despite the LLMs' higher raw accuracy. Key limitations are the single test partition, the absence of formal ethical review, and the five-specialty, single-region scope. Future work should add k-fold evaluation, an ablation separating architecture from input representation, and the fifteen excluded specialties under a formal consent and ethics protocol.

## ACKNOWLEDGEMENTS

The authors sincerely thank Oishi Dhar, Umme Habiba Hima, Tasnova Tasneen, Ami Biswas, Joya Dhar, and Roksana Afrin Anika for generously sharing the dataset they developed for their academic theses and projects, which has been used in this research.

### Conflict of Interest

We certify that there is no conflict of interest with any financial, personal, or other relationships with other people or organization related to the material discussed in the manuscript.

### Declaration of AI Assistance

All zero-shot LLM classification results reported in Section 2.3.1 were produced by the LLMs under evaluation as part of the study's experimental design. The authors reviewed, verified, and edited all AI-assisted content and take full responsibility for the accuracy, originality, and scientific validity of the manuscript.